\documentclass{article}

\usepackage[preprint]{neurips_2026}

\usepackage[utf8]{inputenc} 
\usepackage[T1]{fontenc}    
\usepackage{hyperref}       
\usepackage{url}            
\usepackage{booktabs}       
\usepackage{amsfonts}       
\usepackage{nicefrac}       
\usepackage{microtype}      
\usepackage{xcolor}         

\usepackage{microtype}
\usepackage{graphicx}
\usepackage{subcaption}
\usepackage{booktabs} 
\usepackage{multirow} 
\usepackage[table]{xcolor} 
\usepackage{pifont} 
\definecolor{ourbaseline}{rgb}{0.95,0.95,0.95}
\definecolor{ourgray}{rgb}{0.88,0.88,0.88}
\usepackage{hyperref}
\usepackage{enumitem}
\usepackage{amssymb}
\usepackage[most]{tcolorbox}
\tcbuselibrary{breakable}

\title{Towards Comprehensive Video Captioning via Hierarchical Multi-Dimensional Verifiable Rewards}

\author{%
  \textbf{Weihuang Lin}$^{1,*,\diamond}$ \quad
  \textbf{Xianing Chen}$^{2,*}$ \quad
  \textbf{Zhibin Wang}$^{2,*}$ \quad
  \textbf{Congzhi Zhang}$^{2}$ \\
  \textbf{Yiwei Ma}$^{1}$ \quad
  \textbf{Jiayi Ji}$^{1}$ \quad
  \textbf{Miao Chen}$^{2}$ \quad
  \textbf{Jun Song}$^{2}$ \\
  \textbf{YuCheng}$^{2}$ \quad
  \textbf{Bo Zheng}$^{2}$ \quad
  \textbf{Xiaoshuai Sun}$^{1,\dagger}$ \quad
  \textbf{Rongrong Ji}$^{1}$ \\
  $^{1}$Xiamen University \qquad
  $^{2}$Alibaba Group
}
\begin{document}

\maketitle

{\renewcommand{\thefootnote}{}%
\footnotetext{\noindent\hspace{-1.8em}%
$^{*}$Equal contribution. \quad
$^{\dagger}$Corresponding author. \quad
$^{\diamond}$Work done during an internship at Alibaba.%
}}

\begin{abstract}
Despite significant progress, current models struggle to generate comprehensive video captions, constrained primarily by coarse-grained supervision, which often fails to capture spatial-temporal dynamics and the underlying logical intent.
To address this issue, we introduce MVRCap, a novel framework that reformulates video captioning as a reinforcement learning task steered by structured verification.
Central to our approach is a Fine-Grained Verifiable Reward mechanism, serving as dense supervision to enforce rigorous visual-semantic alignment.
Specifically, we construct this reward system anchored by three complementary pillars: a Hierarchical Concept Coverage Reward utilizing our newly constructed MVRCap-QA dataset to ensure multi-dimensional completeness and semantic depth, a Factuality Verification Reward to enhance factual consistency, and a Conciseness Constraint to maximize semantic density.
Furthermore, we curate the high-quality MVRCap-SFT dataset, providing structured annotations across six key dimensions, to serve as a robust foundational initialization for the RL policy in this training pipeline.
Extensive experiments demonstrate that MVRCap achieves state-of-the-art performance on authoritative benchmarks including VDC, CAPability, and VC4VG. 
%
\end{abstract}

\section{Introduction}

\begin{figure}
  \vspace{-2em}
  \centering
  \includegraphics[width=1.\columnwidth]{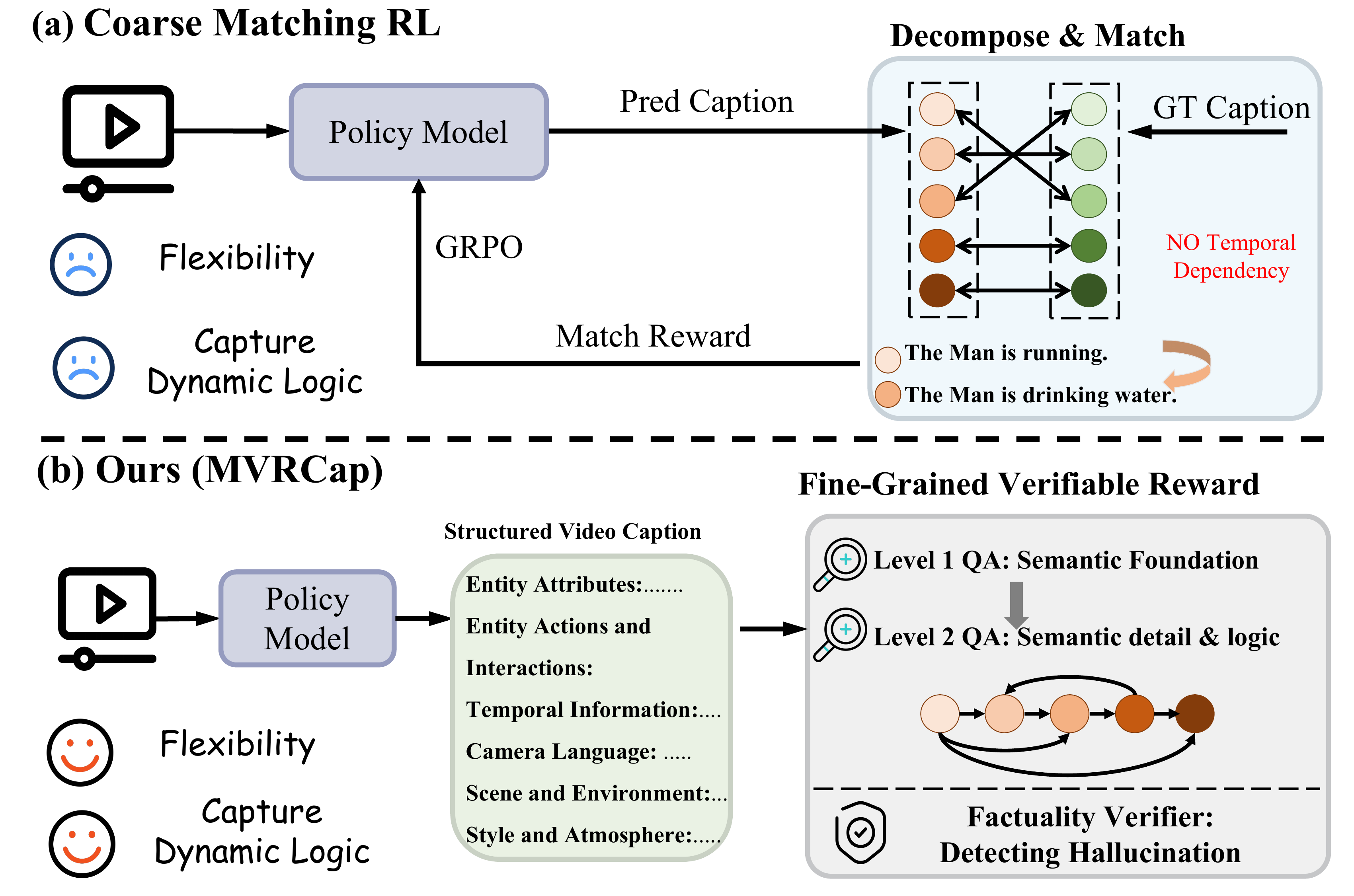}
  \caption{(a) \textbf{Coarse Matching RL} treats captions as unordered sets of atomic facts, failing to capture dynamic logic. (b) In contrast, our approach leverages Structured Visual Probing. We employ a Fine-Grained Verifiable Reward that integrates hierarchical QA probing and factuality verification to enforce supervision across six key dimensions.}
  \label{fig:intro}
\vspace{-2em}
\end{figure}

Video captioning~\cite{chai2024auroracap,islam2024video,lian2025describe,zhou2024streaming} involves translating dynamic visual streams into coherent natural language. Unlike static image captioning~\cite{Xing2025CapRL}, it requires modeling complex temporal dependencies and motion evolution. 

Current approaches~\cite{meng2025videocap,zhong2025owlcap,wang2024tarsier} have achieved remarkable success in identifying salient visual entities and recognizing primary actions. While these models can generate basic inventories of specific events, they often struggle to weave these facts into a comprehensive and logically consistent narrative. Constrained by paradigms that prioritize local object recognition over global scene reasoning, their focus remains tethered to literal physical content. 
This structural limitation results in fundamentally incomplete visual narratives. Specifically, these models frequently overlook the underlying logical intent behind explicit actions and fail to integrate essential observational dynamics, such as viewpoint shifts and lighting, which require holistic perception.

To achieve a comprehensive representation of video content, we propose a holistic captioning taxonomy encompassing six essential dimensions. These dimensions are systematically structured to deconstruct the core facets of visual narratives: \textit{Entity Attributes} and \textit{Scene and Environment} define the \textbf{physical composition}; \textit{Actions and Interactions} and \textit{Temporal Information} model the \textbf{dynamic evolution}; while \textit{Camera Language} and \textit{Style and Atmosphere} determine the \textbf{observational context}.
Given that existing datasets lack caption annotations that comprehensively model video scenes, we leverage advanced commercial MLLMs to establish an automated pipeline, generating high-density, structured captions across these specialized dimensions. Through this pipeline, we curate MVRCap-SFT, a high-quality instruction-tuning dataset with fine-grained multi-dimensional annotations. While Supervised Fine-Tuning on base MLLMs using MVRCap-SFT enables multi-dimensional instruction following, this approach proves insufficient for achieving reliable video understanding. Our analysis reveals that the model prioritizes superficial pattern memorization~\cite{chu2025sft} over genuine visual reasoning. It mimics the linguistic complexity of the synthetic data without robust visual grounding, leading to significant hallucinations, particularly when describing temporal dynamics.


To overcome the disconnect between linguistic fluency and visual grounding, simply applying standard reinforcement learning is insufficient. As illustrated in Fig.~\ref{fig:intro} (a), prevailing Coarse Matching RL approaches treat captions as unordered sets of atomic semantic units~\cite{zhong2025owlcap}. This paradigm fails to capture the temporal dependencies and causal logic essential for comprehensive video understanding. To transcend these limitations, we introduce MVRCap (Fig.~\ref{fig:intro} (b)), a novel framework that enforces genuine visual-semantic alignment via a Fine-Grained Verifiable Reward mechanism.
Specifically, we construct this reward system anchored by three pillars. First, to ensure comprehensive content coverage, we introduce a \textbf{Hierarchical Concept Coverage Reward} underpinned by our curated MVRCap-QA dataset. By actively interrogating the model with multi-level question-answer pairs across six critical dimensions, we transform the captioning objective into a quantifiable metric that probes structural hierarchy and information diversity. Second, to enhance factual consistency, we implement a \textbf{Factuality Verification Reward} as a rigorous guardrail. This mechanism penalizes inconsistencies across objects, actions, and temporal dynamics. Finally, we incorporate a Conciseness Constraint to regularize generation length~\cite{yu2025dapo}, guiding the model to convey maximum semantic richness with minimal redundancy.

Extensive experiments demonstrate MVRCap's superiority, achieving state-of-the-art performance on VDC~\cite{chai2024auroracap}, CAPability~\cite{liu2025capability}, and VC4VG~\cite{du2025vc4vg}. 
The consistent performance gains across benchmarks verify that our approach effectively captures fine-grained visual details while maintaining semantic coherence and structural integrity.

In summary, the contributions of this paper are threefold:

\begin{itemize}[itemsep=0pt, topsep=-3pt]
\item We formulate a comprehensive video captioning schema that systematically deconstructs video content from explicit physical properties to underlying logical semantics. Based on this, we construct the high-quality MVRCap-SFT and MVRCap-QA datasets, providing a rigorous foundational data suite that facilitates the transition of video captioning from flat enumeration to structured visual reasoning.

\item We propose \textbf{MVRCap}, a novel framework that steers models towards rigorous visual fidelity using Reinforcement Learning. We design a \textbf{Hierarchical Multi-Dimensional Verifiable Reward} mechanism, which synergizes Hierarchical Concept Coverage, Factuality Verification, and Conciseness Constraints.

\item Extensive experiments demonstrate that MVRCap significantly outperforms state-of-the-art on benchmarks including VDC, CAPability, and VC4VG.
\end{itemize}

\section{Related Work}

\subsection{Video Captioning}
Video captioning has evolved from generating concise summaries~\cite{yang2023vid2seq,yan2022videococa,hua2025v2xum} to producing comprehensive narratives that fuel Multimodal Large Language Models. A prevailing trend involves leveraging advanced proprietary models to synthesize instruction data for supervised fine-tuning. For instance, ShareGPT4Video~\cite{chen2024sharegpt4video}, AuroraCap~\cite{chai2024auroracap}, and Tarsier~\cite{wang2024tarsier} enhance description density by scaling up detailed annotations of objects and temporal events. More recently, VideoCap-R1~\cite{meng2025videocap} pioneered the integration of reinforcement learning into this domain by employing explicit reasoning steps to analyze subjects and attributes prior to generation.
However, even with reasoning-enhanced strategies, existing methods largely remain confined to a surface-level enumeration paradigm. These approaches tend to treat video understanding as a flat inventory of isolated entities and primary motions, frequently overlooking the underlying causal logic and the holistic structural context of the video. In contrast, our work transcends this literal fact-listing by formulating video captioning as a structured visual reasoning process. By integrating a systematic multi-dimensional taxonomy with a Fine-Grained Verifiable Reward mechanism, MVRCap enforces deep visual-semantic alignment, producing narratives that are not only comprehensively structured but also grounded and logically coherent.

\subsection{Reinforcement Learning with Verifiable Rewards}
Unlike supervised fine-tuning which enforces imitation of specific trajectories, reinforcement learning with verifiable rewards paradigm encourages models to explore diverse solution paths toward an objective and verifiable outcome, achieving remarkable success in linguistics~\cite{shao2024deepseekmath,guo2025deepseek} and multimodal domains~\cite{liu2025visual,feng2025video,wang2025videorft,yang2025longvt,zeng2026video,ding2025videozoomer}. However, applying verifiable rewards to open-ended video captioning remains notoriously difficult due to the lack of objective ground truth.
VideoCap-R1~\cite{meng2025videocap} attempts to use event coverage as a reward, but its metric remains coarse-grained. 
OwlCap~\cite{zhong2025owlcap} advances this by decomposing captions into atomic units for set matching. However, by treating semantic concepts as isolated entities, this paradigm fails to capture the intricate causal logic and temporal dependencies that exist between these atoms. 
In contrast, we propose a reinforcement learning paradigm that shifts from lexical set matching to cognitive verification. Instead of fragmenting captions into isolated units, we utilize hierarchical question-answering to probe the underlying narrative structure and causal logic. This constructs a verifiable reward system that rigorously enforces factual faithfulness.

\begin{figure}[t]
\vspace{-1em}
  \centering
  \includegraphics[width=1\columnwidth]{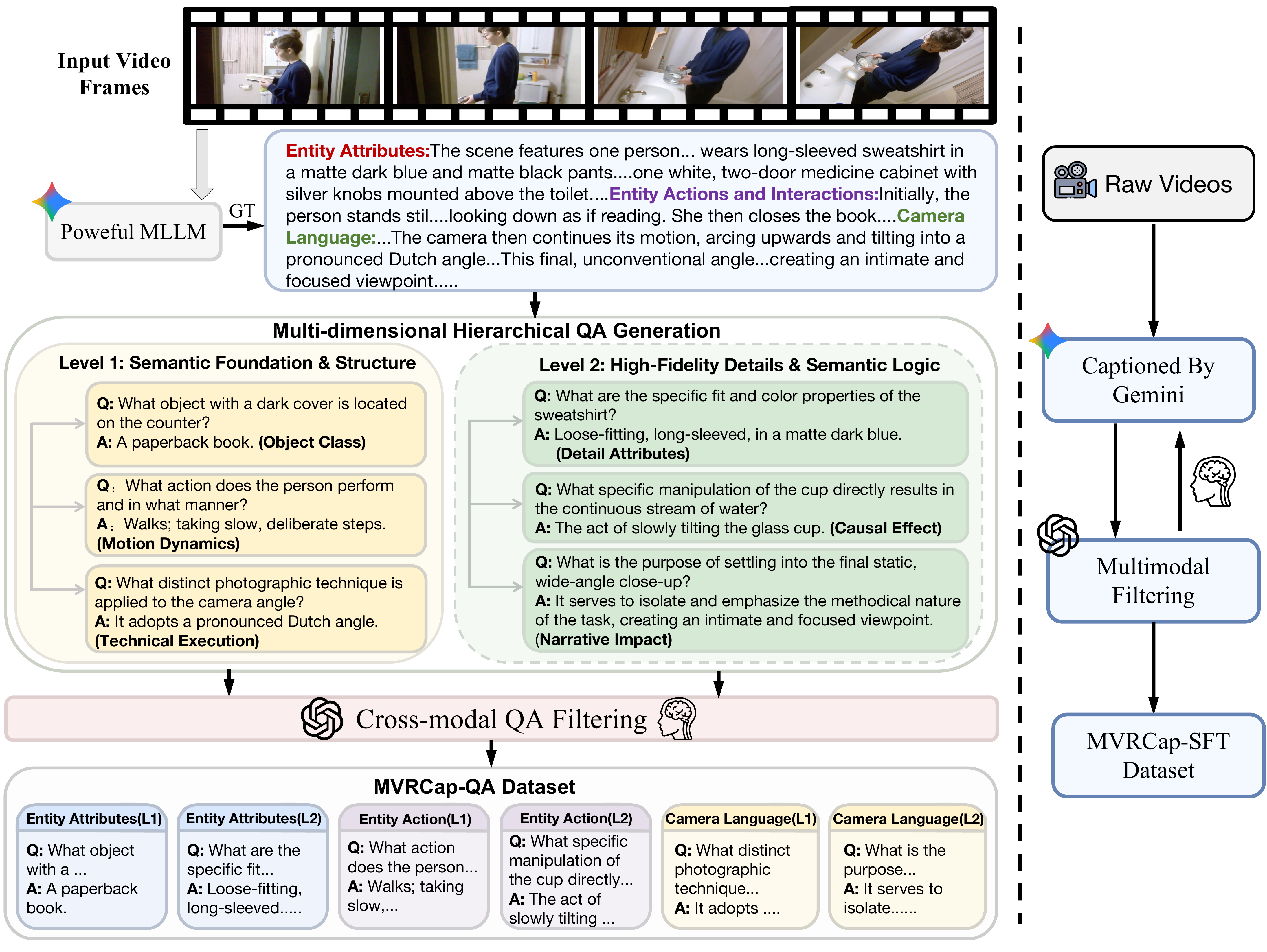}
  \caption{\textbf{The Data Construction Pipeline for MVRCap.} The \textbf{left panel} details the construction of the MVRCap-QA dataset. For brevity, only three dimensions are visualized. The \textbf{right panel} illustrates the workflow for generating the MVRCap-SFT dataset. The human icons denote steps involving manual validation.}
  \label{fig:data}
\vspace{-1em}
\end{figure}

\section{Method}

\subsection{Data Construction Workflow}

To bridge the gap between video content and comprehensive narratives, we establish a systematic pipeline for data synthesis. This workflow transforms subjective video description tasks into a structured semantic parsing objective.
First, we define our Multi-Dimensional Annotation Paradigm to delineate the information requisite for a comprehensive video description in Sec.~\ref{Schema}. 
Second, we leverage a powerful MLLM to generate the MVRCap-SFT dataset by sampling from diverse open-source collections aimed at equipping the model with fundamental multi-dimensional captioning capabilities in Sec.~\ref{sft}. 
Third, we construct the MVRCap-QA dataset to facilitate fine-grained verifiable reward signal in Sec.~\ref{QA}.

\subsubsection{Multi-Dimensional Annotation Paradigm.}
\label{Schema}

To transition from coarse-grained captions to holistic perception, we design a hierarchical paradigm centered on six critical dimensions. This paradigm transforms the task into a structured visual reasoning process. 
\begin{itemize}[itemsep=0pt, topsep=-3pt]
\item \textbf{Entity Attributes.} This dimension mandates the precise identification of subjects and their corresponding detailed attributes. 
\item \textbf{Actions and Interactions.} We focus on the causal logic of motion. Furthermore, we analyze the complex interactions between subjects and their environment, as well as the consequent state transitions within the scene.
\item \textbf{Temporal Information.} This component enforces narrative coherence by reconstructing the chronological evolution of events and scene transitions. 
\item \textbf{Camera Language.} We incorporate cinematic syntax to interpret how technical choices, such as camera movement, focus shift, and depth of field, construct the visual narrative. 
\item \textbf{Scene and Environment.} This dimension concentrates on the physical characteristics of the setting. It details static environmental attributes while tracking dynamic changes throughout the video. 
\item \textbf{Style and Atmosphere.} The final dimension synthesizes the aesthetic tone, explicitly linking visual composition to the underlying mood of the footage. 
\end{itemize}

\subsubsection{MVRCap-SFT Dataset Construction}
\label{sft}
Building upon the proposed paradigm, we establish an automated pipeline to construct the MVRCap-SFT dataset, as shown in the right part of Fig.~\ref{fig:data}. This process leverages the advanced multimodal understanding of Gemini 2.5 Pro~\cite{comanici2025gemini} to scale high-quality annotations. To ensure diversity in visual scenarios, we sample raw videos from open-source datasets, as detailed in Appendix. By integrating the definitions of our Multi-Dimensional Annotation Paradigm Prompt, as detailed in Appendix, directly into the generation instruction, we ensure the model produces dense, logically coherent descriptions that adhere to our structural requirements.

To filter the generated captions and eliminate hallucinations, we employ a Cross-Model Validation approach. Specifically, we use an independent model, GPT-5, to compare the generated captions directly against the source videos, explicitly pruning any hallucinatory content. Furthermore, we introduce human feedback by reviewing a subset of the identified failure cases. Insights from these failure cases are then used to iteratively adjust and optimize the generation prompts. This streamlined process yields 4K video-caption pairs.

\subsubsection{MVRCap-QA Dataset Construction}
\label{QA}

Motivated by DeepSeek-R1's~\cite{guo2025deepseek} success with verifiable reinforcement learning, we adapt this paradigm to video captioning. Yet, unlike math or coding, this task is highly subjective and lacks a definitive ground truth.
To circumvent reward hacking via generic outputs, we transform our 6-D schema evaluation into an open-ended QA assessment. Specifically, we curate the MVRCap-QA dataset using a \textbf{Two-Level Cognitive Hierarchy}. This penalizes superficial recognition and ensures a comprehensive evaluation ranging from basic semantic accuracy to nuanced contextual understanding.

The initial cognitive level, \textbf{Semantic Foundation and Structure}, functions as a mechanism to ensure factual grounding, focusing on the fundamental accuracy of explicit visual content. By interrogating the locations, and basic events within the video, this level aims to prevent generative hallucinations. For static modalities, such as Entity Attributes and Scene, we validate the model's grasp of Entity Presence and Spatial Layout, ensuring the correct perception of subject existence and their arrangement within the environment. In parallel, the assessment of dynamic and technical dimensions centers on Subject Behavioral Performance, Chronological Sequence, and Technical Classification. Consequently, this level confirms the model's foundational capability to identify actions, order events linearly, and classify camera techniques, such as shot sizes, without error.

To transcend basic pattern matching, the second cognitive level, \textbf{High-Fidelity Details and Semantic Logic}, is designed to assess the descriptive depth and causal reasoning underlying the video content. For entities and scenes, we prioritize the richness of visual details and optical atmosphere, requiring the model to capture fine-grained attributes. Regarding action-oriented and stylistic dimensions, the assessment concentrates on Interaction Logic, Environmental Impact, and Aesthetic Intent. Rather than simply cataloging isolated events, the model must explain how actions physically transform their surroundings, the narrative coherence governing temporal transitions, and the manner in which specific lighting or cinematographic choices actively construct the video's atmospheric mood. Please refer to the Appendix for more details.

By systematically extracting QA pairs across these two levels for each dimension, we processed another 5K video samples, as detailed in Appendix, yielding a total of 205K QA pairs. This transformation converts the subjective captioning task into a verifiable matrix of objective facts and logical deductions, providing a robust supervision signal for reinforcement learning. 
First, we utilize GPT-5 to verify the extracted QA pairs directly against the source videos, pruning or correcting any ungrounded instances. Crucially, human experts not only review failure cases to iteratively refine the extraction prompts, but also conduct final manual quality assurance on randomly sampled batches of the finalized dataset. This rigorous, visually grounded curation ensures that all QA pairs are anchored in visual evidence rather than linguistic artifacts.

\begin{figure}
  \vspace{-2em}
  \centering
  \includegraphics[width=1\columnwidth]{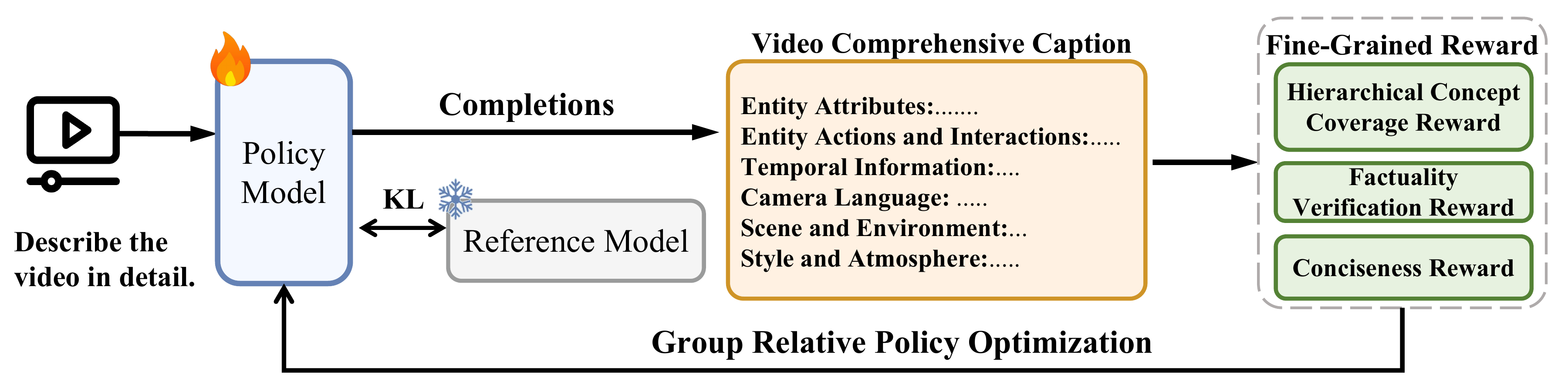}
  \caption{\textbf{The Overall Framework of MVRCap.}}
  \label{fig:method}
\vspace{-2em}
\end{figure}

\subsection{Training Strategy}

Our training framework proceeds in two distinct phases. We first establish a foundational capability for multi-dimensional instruction following via Supervised Fine-Tuning, followed by a reinforcement learning stage that enforces rigorous visual-semantic alignment using GRPO.

\subsubsection{Supervised Fine-Tuning}
We initialize our model by fine-tuning a base MLLM on the MVRCap-SFT dataset to adapt it to our multi-dimensional schema and establish preliminary visual-semantic alignment.  Given a video input $V$ and the corresponding ground-truth structured caption $Y$, the loss function is 
$\mathcal{L}_{SFT} = - \sum_{t=1}^{T} \log \pi_\theta(y_t | V, y_{<t})$,
where $\pi_\theta$ represents the policy of the model parameterized by $\theta$, $y_t$ denotes the token at time step $t$, and $y_{<t}$ represents the preceding context. This stage ensures that the model learns to generate fluent and structured text, providing a stable reference policy $\pi_{ref}$ for the subsequent reinforcement learning phase.

\subsubsection{Group Relative Policy Optimization}
While SFT provides formatted outputs, it lacks a mechanism to penalize hallucinations or enforce deep visual grounding. To address this, we employ Group Relative Policy Optimization (GRPO)~\cite{team2025kimi}.
For each video query $q$, GRPO samples a group of $G$ outputs $\{o_1, o_2, ..., o_G\}$ from the current policy $\pi_{\theta_{old}}$. For each output $o_i$, we compute a composite reward $r_i$. The advantage $A_i$ for each output is calculated by standardizing the rewards within the group:
$A_i = \frac{r_i - \text{mean}(\mathbf{r})}{\text{std}(\mathbf{r}) + \epsilon}$,
where $\mathbf{r} = \{r_1, ..., r_G\}$ represents the vector of rewards for the sampled group.

The training objective seeks to maximize the surrogate loss function, which incorporates a clipping mechanism to prevent destructive policy updates and a regularization term to maintain stability:

\begin{equation}
\mathcal{L}_{GRPO}(\theta) = \mathbb{E}_{q \sim P(Q), \{o_i\} \sim \pi_{\theta_{old}}} \left[ \frac{1}{G} \sum_{i=1}^G \left( \min \left( \rho_i A_i, \text{clip}(\rho_i, 1-\epsilon, 1+\epsilon) A_i \right) - \beta \mathbb{D}_{KL}(\pi_\theta || \pi_{ref}) \right) \right]
\end{equation}
where $\rho_i = \frac{\pi_\theta(o_i|q)}{\pi_{\theta_{old}}(o_i|q)}$ denotes the probability ratio between the new and old policies, $\text{clip}(\cdot)$ restricts the ratio to the range $[1-\epsilon, 1+\epsilon]$, and $\mathbb{D}_{KL}(\pi_\theta || \pi_{ref})$ represents the KL divergence between the current policy $\pi_\theta$ and the reference SFT model $\pi_{ref}$.

\subsubsection{Fine-Grained Verifiable Reward Formulation}
The efficacy of GRPO relies heavily on the reward signal $r_i$. To enforce strict visual grounding and logical coherence, we formulate a composite reward comprising three pillars: a \textbf{Hierarchical Concept Coverage Reward}, a \textbf{Factuality Verification Reward}, and a \textbf{Conciseness Constraint}.

\noindent\textbf{Hierarchical Concept Coverage Reward.}

An ideal caption serves as a standalone semantic proxy for the visual content, negating the need for a viewer to revisit the source video to extract information. Consequently, we formulate the coverage reward $R_{cov}$ based on the principle of Question Answering capability. We utilize the curated MVRCap-QA dataset, which consists of hierarchical question-answer pairs derived from the ground truth. Specifically, for each video sample, we meticulously organized the corresponding QA pairs across six critical dimensions, with each dimension further stratified into specific hierarchical levels.

For a generated caption $C_{pred}$, we employ a frozen Judge LLM $\mathcal{J}$ to evaluate whether $C_{pred}$ contains sufficient information to correctly answer a set of dimension-specific questions. Importantly, the Judge determines the answer solely based on $C_{pred}$ without accessing the video. The detailed prompting schema for Judge is provided in Appendix. To ensure balanced optimization across all semantic aspects, we adopt a dimension-wise aggregation strategy. We first compute the accuracy within each specific dimension $d$ and then average these dimension-level scores to obtain the final reward. Formally, let $\mathcal{D}$ be the set of all defined dimensions and $Q_d$ be the set of questions belonging to dimension $d$. The Concept Coverage Reward $R_{cov}$ is calculated as:

\begin{equation}
R_{cov} = \frac{1}{|\mathcal{D}|} \sum_{d \in \mathcal{D}} \underbrace{\left( \frac{1}{|Q_d|} \sum_{q \in Q_d} \mathbb{I}\left( \mathcal{J}(q, C_{pred}) \equiv A_{gt} \right) \right)}_{R_d},
\end{equation}

where $\mathbb{I}(\cdot)$ is the indicator function, $A_{gt}$ is the ground truth answer, and $R_d$ represents the coverage score for dimension $d$. This compels the model to adopt a strategy of maximal semantic coverage, effectively learning to intrinsically embed deep visual reasoning and evidence into its descriptions to satisfy potential queries across all cognitive levels.

\noindent\textbf{Factuality Verification Reward.}
While coverage ensures comprehensiveness, it does not guarantee accuracy. Video captioning is particularly susceptible to specific types of hallucinations that compromise the narrative integrity. To suppress these errors, we introduce a Factuality Verification Reward $R_{fact}$. Specifically, we leverage the Judge LLM to scrutinize the predicted caption $C_{pred}$ against the ground truth $C_{gt}$, specifically targeting the three high-risk failure modes most prevalent in video descriptions: (1) \textit{Object Hallucination}, which penalizes the fabrication of non-existent entities; (2) \textit{Action Hallucination}, which detects contradictions in motion dynamics or interactions; and (3) \textit{Temporal Hallucination}, which identifies disorders in the chronological sequence relative to the ground truth. 
The detailed prompting schema is illustrated in Appendix. Based on these strict criteria, \textbf{the Judge assigns a graded scalar score $R_{fact}$} that is inversely proportional to the severity of detected errors; effectively, a \textbf{higher score indicates fewer contradictions} and higher factual consistency. Crucially, by instructing the Judge to detect contradictions rather than measuring surface similarity, we penalize factual errors without punishing valid paraphrases. This distinction is vital: it ensures that the optimization objective is semantic consistency rather than lexical replication. Consequently, the model is incentivized to align with the ground truth's logic while retaining the freedom to utilize diverse syntactic structures and synonyms.

\noindent\textbf{Conciseness Constraint.}
Maximizing coverage can inadvertently encourage verbosity. To address this, we introduce a Concise Constraint $R_{len}$ to promote high information density. We implement a soft punishment mechanism that applies a linear penalty only when the generated length exceeds a target threshold $L_{target}$. This approach grants the model flexibility for complex descriptions while effectively penalizing unnecessary redundancy.

\noindent\textbf{Total Reward Formulation.}
The final reward $r_i$ for a given video-caption pair is formulated as the weighted sum of the three holistic components:
\begin{equation}
r_i = \lambda_{cov} R_{cov} + \lambda_{fact} R_{fact} + \lambda_{len} R_{len},
\end{equation}
where $R_{len} = \max\left(-1, \min\left(0, \frac{(L_{max} - L_{cache}) - \text{len}(C_{pred})}{L_{cache}}\right)\right)$ represents the length penalty term.

\section{Experiments}

\begin{table}[!t]
    \centering
    \caption{Quantitative captioning evaluation results on VC4VG-Bench.}
    \label{tab:vc4vg}
    
    \small 
    \setlength{\tabcolsep}{1pt} 
    \renewcommand{\arraystretch}{1.1} 
    
    \begin{tabular}{lcccccccc} 
    \toprule
    \multirow{2}{*}{\textbf{Model}}  &Environment & Subject & Motion & Camera & Style & Necessity-L1 & Necessity-L2 &Total score \\
& Score/\% & Score/\% & Score/\% & Score/\% & Score/\% & Score/\% & Score/\%& Score/\%\\
    \midrule    ShareCaptioner~\cite{chen2024sharegpt4video} & 196/43.5 & 103/22.3 & 85/25.4 & 48/33.1 & 12/70.6 & 284/46.3 & 160/20.1 & 444/31.5 \\
    Vriptor~\cite{yang2024vript} & 208/46.1 & 126/27.3 & 60/17.9 & 31/21.4 & 16/94.1 & 303/49.3 & 138/17.3 & 441/31.3 \\
    VideoLLaMA3~\cite{zhang2025videollama} & 119/26.4 & 106/22.9 & 88/26.3 & 17/11.7 & 14/82.4 & 232/37.8 & 112/14.1 & 344/24.4 \\
    CogVLM2Cap~\cite{yang2024cogvideox} & 216/47.9 & 174/37.7 & 93/27.8 & 14/9.7 & 13/76.5 & 317/51.6 & 193/24.2 & 510/36.2 \\
    Tarsier2~\cite{yuan2025tarsier2} & 170/37.7 & 155/33.5 & 164/49.0 & 29/20.0 & 16/94.1 & 308/50.2 & 226/28.4 & 534/37.9 \\
    LLaVA-Video-Gen~\cite{du2025vc4vg} & 304/67.4 & 256/55.4 & 154/46.0 & 74/51.0 & 16/94.1 & 459/74.8 & 345/43.3 & 804/57.0 \\

    \midrule

    Gemini-1.5-Pro & 372/82.5 & 328/71.0 & 170/50.7 & 85/58.6 & 17/100.0 & 513/83.6 & 459/57.7 & 972/68.9 \\
    Gemini-2.5-Pro & 368/81.6 & 319/69.1 & 230/68.7 & 85/58.6 & 17/100.0 & 504/82.1 & \underline{515/64.7}& \underline{1019/72.3} \\
    Gemini-3.1-Pro   & 359/79.6 & 339/73.4 & 199/59.4 & 91/62.8 & 16/94.1  & 507/82.6 & 497/62.4 & 1004/71.2 \\
    
    \midrule
    
    \rowcolor[gray]{0.95} Qwen3-VL-8B~\cite{bai2025qwen2} & 269/59.7 & 181/39.2 & 107/31.9 & 28/19.3 & 16/94.1 & 357/58.1 & 244/30.7 & 601/42.6 \\
    \rowcolor[gray]{0.95} \textbf{MVRCap-8B (Ours)} & 348/77.2 & 295/63.9 & 182/54.3 & 80/55.2 & 17/100.0 & 497/80.9 & 425/53.4 & 922/65.4 \\
    
    \midrule
    
    \rowcolor[gray]{0.95} Qwen3-VL-32B~\cite{bai2025qwen3} & 341/75.6 & 311/67.3 & 205/61.2 & 59/40.6 & 17/100.0 & 460/74.9 & 473/59.4 & 933/66.2 \\
    \rowcolor[gray]{0.9} \textbf{MVRCap-32B (Ours)} & \textbf{379/84.0} & \textbf{353/76.4} & \textbf{222/66.3} & \textbf{81/55.9} & \textbf{17/100.0} & \textbf{513/83.6} & \textbf{539/67.7} & \textbf{1052/74.6}\\

    \bottomrule
    \end{tabular}
\end{table}

\begin{table*}[!t] 
\setlength\tabcolsep{2.5pt}
\centering
\caption{Quantitative comparison on the CAPability benchmark. The F1 Score across all dimensions. ``Video Avg.'' represents the mean score of the four dynamic dimensions.}
\label{tab:Cap_f1score}
\vspace{-0.5em}
\resizebox{\textwidth}{!}{%
\begin{tabular}{@{}llccccccccccccccc@{}}
\toprule
 & \textbf{Model} 
 & \textbf{\begin{tabular}[c]{@{}c@{}}Obj.\\ Cate.\end{tabular}} 
 & \textbf{\begin{tabular}[c]{@{}c@{}}Obj.\\ Num.\end{tabular}} 
 & \textbf{\begin{tabular}[c]{@{}c@{}}Obj.\\ Color\end{tabular}} 
 & \textbf{\begin{tabular}[c]{@{}c@{}}Spa.\\ Rel.\end{tabular}} 
 & \textbf{Scene} 
 & \textbf{\begin{tabular}[c]{@{}c@{}}Cam.\\ Ang.\end{tabular}} 
 & \textbf{OCR} 
 & \textbf{Style} 
 & \textbf{\begin{tabular}[c]{@{}c@{}}Cha.\\ Iden.\end{tabular}} 
 & \textbf{\begin{tabular}[c]{@{}c@{}}(D) Obj.\\ Num.\end{tabular}} 
 & \textbf{Act.} 
 & \textbf{\begin{tabular}[c]{@{}c@{}}Cam.\\ Mov.\end{tabular}} 
 & \textbf{Event} 
 & \textbf{\begin{tabular}[c]{@{}c@{}}Video\\ Avg.\end{tabular}} 
 & \textbf{\begin{tabular}[c]{@{}c@{}}All\\ Avg.\end{tabular}} \\ \midrule 
& InternVL2.5-8B & 69.01 & 31.08 & 60.78 & 54.18 & 75.23 & 42.72 & 77.62 & 76.10 & 12.74 & 16.12 & 30.86 & 18.44 & 42.11 & 26.88 & 46.69 \\
& LLaVA-OV-7B & 71.90 & 28.66 & 61.40 & 53.98 & 75.66 & 41.76 & 70.29 & 74.80 & 6.81 & 17.07 & 32.31 & 7.69 & 40.42 & 24.37 & 44.83 \\
& NVILA-8B & 73.43 & 34.21 & 63.30 & 52.22 & 72.64 & 41.29 & 75.34 & 72.70 & 9.89 & 9.06 & 22.25 & 7.28 & 31.19 & 17.45 & 43.45 \\
& VideoLLaMA3-7B & 72.85 & 31.80 & 65.64 & 59.51 & 78.33 & 41.63 & 74.41 & 73.04 & 4.86 & 7.32 & 34.05 & 25.03 & 44.31 & 27.68 & 47.14 \\
& VideoChat-R1-7B & 76.85 & 30.21 & 64.94 & 55.56 & 80.64 & 45.87 & 80.19 & 71.28 & 12.53 & 18.97 & 29.66 & 22.76 & 38.61 & 27.50 & 48.31 \\ \cmidrule{2-17}
\rowcolor[gray]{0.95} & Qwen2.5-VL-7B & 75.72 & 27.62 & 65.59 & 56.73 & 80.56 & 47.38 & 81.49 & 72.70 & 12.14 & 16.95 & 37.47 & 26.02 & 44.09 & \underline{31.13} & \underline{49.57} \\ 
\rowcolor[gray]{0.9} & \textbf{MVRCap} & \textbf{81.16} & \textbf{40.39} & \textbf{74.08} & \textbf{66.50} & \textbf{82.09} & \textbf{51.29} & \textbf{82.13} & \textbf{70.20} & \textbf{20.30} & \textbf{19.60} & \textbf{45.03} & \textbf{34.63} & \textbf{49.89} & \textbf{37.29} & \textbf{55.18} \\ \bottomrule
\end{tabular}%
}
\end{table*}

\begin{table}[!t]
    \centering
    \caption{Quantitative comparison on the VDC benchmark.}
    \label{tab:vdc}
    \small %
    \setlength{\tabcolsep}{3pt} %

    \begin{tabular}{lcccccc} 
    \toprule
    \multirow{2.5}{*}{\textbf{Model}} & \textbf{Detailed} & \textbf{Camera} & \textbf{Short} & \textbf{Background} & \textbf{Object} & \textbf{Average} \\
    & (Acc/Score) & (Acc/Score) & (Acc/Score) & (Acc/Score) & (Acc/Score) & (Acc/Score) \\
    \midrule
    \multicolumn{7}{l}{\textit{\textbf{General MLLM}}} \\
    VILA-v1.5-8B~\cite{lin2024vila} & 42.8/2.2 & 39.7/2.1 & 39.3/2.0 & 39.8/2.1 & 40.9/2.1 & 40.5/2.1 \\
    VideoChat2-7B~\cite{li2024mvbench} & 40.5/2.1 & 31.9/1.7 & 40.2/2.1 & 34.9/1.8 & 34.9/1.8 & 36.5/1.9 \\
    InternVL2-8B~\cite{chen2024expanding} & 34.9/1.8 & 39.1/2.1 & 33.0/1.7 & 37.5/1.9 & 44.2/2.2 & 33.7/2.0 \\
    LLaVA-OneVision-7B~\cite{li2024llava} & 41.8/2.2 & 37.6/2.0 & 41.6/2.1 & 34.3/1.8 & 38.8/2.0 & 38.8/2.0 \\
    LLaVA-Video-7B~\cite{zhang2024video} & 35.0/1.8 & \underline{46.1/2.3} & 32.8/1.7 & 37.6/1.9 & 46.2/2.4 & 39.0/2.0 \\
    Video-R1-7B~\cite{feng2025video} & \underline{45.6/2.4} & 42.7/2.2 & \underline{44.5/2.3} & 40.6/2.1 & 45.9/2.3 & 43.9/2.3 \\
    \midrule
    \multicolumn{7}{l}{\textit{\textbf{Video Caption MLLMs}}} \\
    ShareGPT4Video-8B~\cite{chen2024sharegpt4video} & 35.6/1.8 & 33.3/1.8 & 39.1/1.9 & 35.8/1.8 & 37.1/1.9 & 36.2/1.9 \\
    Vriptor~\cite{yang2024vript} & 38.5/2.0 & 37.6/2.0 & 38.4/2.0 & 37.1/1.9 & 37.0/1.9 & 37.7/2.0 \\
    AuroraCap-7B~\cite{chai2024auroracap} & 41.3/2.1 & 43.5/2.3 & 32.1/1.7 & 35.9/1.8 & 39.0/2.0 & 38.2/2.0 \\
    Tarsier-7B~\cite{wang2024tarsier} & 38.3/2.1 & 42.6/2.3 & 41.7/2.2 & 36.4/1.9 & 42.2/2.2 & 40.2/2.1 \\
    VideoCap-R1-7B~\cite{meng2025videocap} & 43.8/2.4 & 41.7/2.3 & 35.2/1.9 & \underline{47.2/2.5} & 47.0/2.5 & 43.0/2.3 \\
    \midrule
     \rowcolor[gray]{0.95} Qwen2.5-VL-7B~\cite{bai2025qwen2} & 46.1/2.3 & 41.3/2.2 & 37.5/2.0 & 45.8/2.4 & 46.9/2.4 & 43.5/2.2 \\
    \rowcolor[gray]{0.9} \textbf{MVRCap (Ours)} & \textbf{51.6/2.7} & \textbf{49.2/2.6} & 41.4/2.2 & \textbf{53.6/2.8} & \textbf{52.5/2.7} & \textbf{49.7/2.6} \\
    \bottomrule
    \end{tabular}
    \vspace{-1em}
\end{table}

\subsection{Implementation Details.}
We employ Qwen2.5-VL-7B-Instruct~\cite{bai2025qwen2} as the default baseline model. The training process consists of two stages: SFT and RL. 
During the SFT phase, we use the Adam optimizer with a learning rate of $1 \times 10^{-5}$. In the RL phase, we set a default frame rate of 2.0 FPS. The values for $L_{max}$ and $L_{cache}$ are set to 1280 and 256. We set the reward weight to 1.0. The RL process is configured with 8 rollouts per iteration, utilizing Qwen3-30B-A3B-Instruct~\cite{yang2025qwen3} as the judge model. All experiments are conducted on 16 NVIDIA H100 GPUs. Refer to the Appendix for further implementation and training specifics.

\subsection{Main Results}
We evaluate our model using three mainstream benchmarks: VC4VG~\cite{du2025vc4vg}, CAPability~\cite{liu2025capability}, and VDC~\cite{chai2024auroracap}. Refer to Appendix for detailed descriptions. In this section, we report the competitive performance achieved by our model across these three benchmarks.

Additionally, human blind evaluations are included in Appendix to further demonstrate the superior captioning quality of our method.

\noindent\textbf{Results on VC4VG}
Table~\ref{tab:vc4vg} evaluates caption quality on the VC4VG benchmark. MVRCap-32B achieves a Total Score of 74.6\%, significantly outperforming LLaVA-Video-Gen (57.0\%) and remaining competitive with proprietary Gemini models. Even our 7B model (65.4\%) yields substantial gains over its Qwen2.5-VL baseline (42.6\%). Consistently high scores across dynamic precision and Necessity-L1/L2 demonstrate that MVRCap generates highly accurate and structurally comprehensive video narratives.

\begin{table}[t]
    \centering
    \small

    \begin{minipage}[t]{0.52\textwidth} 
        \centering
        \caption{Ablation study on the effectiveness of different reward components. $R_{cov}$, $R_{fact}$, and $R_{len}$ denote Coverage, Factuality, and Conciseness respectively.}
        \label{tab:ab_reward}
        \setlength{\tabcolsep}{3pt} 
        \begin{tabular}{ccccccc}
        \toprule
        \multicolumn{3}{c}{Components} & \multicolumn{2}{c}{VDC} & \multicolumn{2}{c}{Capability} \\
        \cmidrule(lr){1-3} \cmidrule(lr){4-5} \cmidrule(lr){6-7}
        $R_{cov}$ & $R_{fact}$ & $R_{len}$ & Acc & Score & Video F1 & All F1 \\
        \midrule
        & & & 45.23 & 2.38 & 30.95 & 50.46 \\
         & \checkmark & \checkmark & 47.12 & 2.42 & 33.77 & 52.85 \\
        \checkmark & & \checkmark & 47.85 & 2.45 & 34.12 & 53.21 \\
        \checkmark & \checkmark & & 48.72 & 2.52 & 35.84 & 54.45 \\
        \checkmark & \checkmark & \checkmark & \textbf{49.67} & \textbf{2.60} & \textbf{37.29} & \textbf{55.18} \\
       
        \bottomrule
        \end{tabular}
    \end{minipage}
    \hfill 
    \begin{minipage}[t]{0.45\textwidth}
        \centering
        \caption{Ablation study on different training stages. We compare the \textit{Base} Qwen2.5-VL-7B with combinations of Supervised Fine-Tuning and Reinforcement Learning.}
        \label{tab:stage_ablation}
        \vspace{1em}
        \setlength{\tabcolsep}{2pt}
        \begin{tabular}{lcccc}
        \toprule
        \multirow{2}{*}{Method} & \multicolumn{2}{c}{VDC} & \multicolumn{2}{c}{Capability} \\
        \cmidrule(lr){2-3} \cmidrule(lr){4-5}
         & Acc & Score & Video F1 & All F1 \\
        \midrule
        Base & 43.53 & 2.24 & 31.13 & 49.57 \\
        Base + SFT & 45.23 & 2.38 & 30.95 & 50.46 \\
        Base + RL & 44.28 & 2.31 & 32.24 & 51.64 \\
        Base + SFT + RL & \textbf{49.67} & \textbf{2.60} & \textbf{37.29} & \textbf{55.18} \\
        \bottomrule
        \end{tabular}
    \end{minipage}
\vspace{-1em}    
\end{table}

\noindent\textbf{Results on CAPability}
Table~\ref{tab:Cap_f1score} reports F1 scores on the CAPability benchmark, where MVRCap achieves state-of-the-art performance among 7B open-source models. Notably, it attains an \textit{All Avg.} of \textbf{55.18}, outperforming the strong Qwen2.5-VL baseline by \textbf{5.61} points. Improvements are particularly pronounced in dynamic dimensions (\textit{Video Avg.} of \textbf{37.29}), demonstrating superior temporal reasoning in \textit{Action} (+7.56) and \textit{Camera Movement} (+8.61). Furthermore, significant gains in static dimensions (e.g., \textit{Object Number}, \textit{Spatial Relations}) confirm that our fine-grained reward mechanism effectively enhances both fundamental visual perception and complex temporal reasoning.

\noindent\textbf{Results on VDC}
Table~\ref{tab:vdc} presents results on the VDC benchmark, where MVRCap outperforms both general and caption-specific MLLMs. Setting a new state-of-the-art, it achieves an Average Accuracy of 49.7\% and a Score of 2.6, significantly surpassing the strong Qwen2.5-VL baseline. Across fine-grained dimensions, MVRCap dominates four of the five categories, notably excelling in \textit{Background} (53.6\%) and \textit{Object} (52.5\%). This confirms our hierarchical verifiable reward effectively enhances dense detail capture and minimizes hallucinations. Furthermore, MVRCap establishes a clear lead over the competitive OwlCap-7B in \textit{Detailed} descriptions, validating its ability to generate comprehensive, logically coherent long-form narratives.

\subsection{Ablation Study}

\noindent\textbf{Synergy of Reward Components.} 
Table~\ref{tab:ab_reward} investigates the contribution of each individual component within our reward mechanism. We observe that removing $R_{cov}$ results in the most significant performance decline where the VDC Score drops remarkably from 2.60 to 2.42. This indicates its pivotal role in guiding the model to capture comprehensive semantic details. Furthermore the omission of $R_{fact}$ leads to a distinct decrease in the Capability Video F1 from 37.29 to 34.12. This suggests that strict factual verification is indispensable for precise dynamic event understanding. The optimal results are achieved only when $R_{cov}$, $R_{fact}$, and $R_{len}$ function synergistically to reach the peak All F1 score of 55.18.

\noindent\textbf{Role of SFT Initialization.} 

Table~\ref{tab:stage_ablation} validates the necessity of our two-stage training pipeline. We observe that applying reinforcement learning directly to the baseline model yields suboptimal results compared to the full two-stage approach. This performance gap indicates that pure RL suffers from a cold-start problem where the model lacks the fundamental schema for comprehensive video description. Without the supervised fine-tuning phase to instill the ability to cover multi-dimensional semantics, the RL optimization struggles to effectively guide the policy. Consequently, the combination of SFT and RL is essential. The SFT stage establishes a robust structural foundation while the RL stage further refines visual grounding and information density to reach SOTA.

\section{Conclusion}
In this work, we introduced MVRCap, a framework for comprehensive video captioning. We constructed two complementary datasets: \textbf{MVRCap-SFT} for establishing a semantic foundation and \textbf{MVRCap-QA} for fine-grained policy optimization. Furthermore, we designed a Fine-Grained Verifiable Reward that explicitly enforces both hierarchical concept coverage and precise factuality.

\bibliographystyle{plain} 
{
	\small
	\bibliography{egbib}
}

\end{document}